\documentclass{article}
\usepackage{ijcai22}

\usepackage{times}
\usepackage{soul}
\usepackage{url}
\usepackage[hidelinks]{hyperref}
\usepackage[utf8]{inputenc}
\usepackage[small]{caption}
\usepackage{graphicx}
\usepackage{amsmath}
\usepackage{amsthm}
\usepackage{amssymb}
\usepackage{booktabs}
\usepackage{algorithm}
\usepackage{algorithmic}
\title{Metric Policy Representations for Opponent Modeling}

\author{
Haobin Jiang
\quad
Yifan Yu \quad  
Zongqing Lu 
\affiliations
\emails
Peking University
}

\begin{document}

\maketitle

\begin{abstract}
In multi-agent reinforcement learning, the inherent non-stationarity of the environment caused by other agents’ actions posed significant difficulties for an agent to learn a good policy independently. One way to deal with non-stationarity is opponent modeling, by which the agent takes into consideration the influence of other agents’ policies. Most existing work relies on predicting other agents’ actions or goals, or discriminating between different policies. However, such modeling fails to capture the similarities and differences between policies simultaneously and thus cannot provide enough useful information when generalizing to unseen agents. To address this, we propose a general method to learn representations of other agents’ policies, such that the distance between policies is deliberately reflected by the distance between representations, while the policy distance is inferred from the sampled joint action distributions during training. We empirically show that the agent conditioned on the learned policy representation can well generalize to unseen agents in three multi-agent tasks.
\end{abstract}

\section{Introduction}

In recent years, deep reinforcement learning (RL) achieved tremendous success in a range of complex tasks, such as Atari games \cite{mnih2015DQN}, Go \cite{silver2017mastering}, and StarCraft \cite{vinyals2019grandmaster}. However, real-world scenarios often requires multiple agents instead of one. With the introduction of other agents, the environment is no longer stationary in the view of each individual agent in the multi-agent system, when the joint policy of other agents is changing. The non-stationary nature and the explosion of dimensions pose many challenges to learning in multi-agent environments.

To address these challenges, centralized training \cite{lowe2017multi,foerster2018counterfactual}, communication \cite{foerster2016learning,sukhbaatar2016learning,jiang2018learning}, value decomposition \cite{sunehag2018value,rashid2018qmix,wang2020qplex,zhang2021fop} and opponent modeling \cite{he2016opponent,hong2018DPIQN,raileanu2018modeling} are proposed successively in attempts to improve the performance of deep RL algorithms in various multi-agent settings, ranging from fully cooperative games to zero-sum games. 

One way to restrain non-stationarity is to distinguish between the invariant dynamics of the environment and the influence of other agents' joint policy, and consider them separately in order to learn an effective policy. In this way, opponent modeling (also termed as agent modeling) has become one of the main research directions, in which a model is trained to predict or represent some properties of other agents, such as goals or actions \cite{albrecht2018autonomous}. Thus, the agent's decision takes into account the dynamics of the environment and the predicted properties of other agents separately, leading to improved performance during training and execution.

In this paper, we focus on the multi-agent learning problem that one \textit{ego} agent learns while interacting with other agents (collectively termed as \textit{opponents} for convenience, regardless of whether they are competitive or not), whose policies are sampled from a set of fixed policies at the beginning of each episode. To perform well, the ego agent should be able to distinguish different opponents' policies and adopt its corresponding policy. More importantly, we expect the ego agent to be immediately generalizable, \textit{i.e.}, to adapt quickly and achieve high performance when facing unseen opponents during execution without updating parameters. Note that this setting is different from the continuous adaptation problem \cite{al2017continuous,kim2020policy}, where both the ego agent and opponents learn continuously.

Similarity can be defined as the distance in physical space or in mental space \cite{shepard1957stimulus}, and plays an important role in problem solving, reasoning, social decision making, \textit{etc.} \cite{HAHN20031}. Some cognitive and social science theories suggest that the observer constructs mental representations for persons. When encountering a new person, the observer makes judgements and inferences based on the similarity between the new individual and known ones \cite{smith1992exemplar}. Therefore, we believe it is also important to consider the similarity between opponents' policies when modeling them, rather than mere distinction. Inspired by this theory, we propose \textit{Metric Policy Representations} (MPR), a novel method to learn policy representations that reflect both similarities and differences by capturing the distances between opponents' policies.

In multi-agent tasks, for the ego agent, the distance between opponents' policies are naturally reflected in the difference between action patterns, and eventually in that of joint-action distributions that can be sampled in interactions with different policies. MPR exploits these distributions to quantify the policy distance so as to essentially model the policy space, and embeds the distance information in the corresponding policy representations. In this way, MPR can accomplish quick adaptation and generalization no matter in competitive or cooperative, partially or fully observable settings. Through experiments, we demonstrate that MPR can greatly improve the learning of existing RL algorithms, especially when interacting with opponents with unseen policies. We further show that the learned policy representations correctly reflect the relations between policies.

Our main contributions are as follows:
\begin{itemize}
    \vspace{-0.1cm}
    \setlength\itemsep{0em}
    \item We propose a metric learning method to learn policy representations that embed policy distance.
    \item We exploit distance between joint action distributions for both discrete and continuous actions to estimate distance between opponent's policies.
    \item We show the performance improvement of MPR combined with DQN \cite{mnih2015DQN} and PPO \cite{DBLP:journals/corr/SchulmanWDRK17} in three multi-agent environments, including Google Research Football \cite{kurach2020google}.
\end{itemize}

\section{Related Work}

In continuous adaptation, the non-stationarity comes from the update of opponents' policies. LOLA \cite{foerster2018learning} takes opponents’ learning process into consideration, where the agent acquires high rewards by shaping the learning directions of the opponents. Al-Shedivat \textit{et al.} [\citeyear{al2017continuous}] proposed a meta-learning method 
and Kim \textit{et al.} [\citeyear{kim2020policy}] extended this method by introducing the opponent learning gradient. 

Unlike continuous adaptation, in settings like ours where the opponents re-select a policy between episodes, direct modeling of opponents becomes effective. Policy reuse methods \cite{hernandez2016identifying,zheng2018deep} and SAM \cite{everett2018learning} maintain a library of opponent models and a library of response policies to learn against non-stationary opponents. When facing unseen opponents, the ego agent can learn a new opponent model or calibrate existing opponent models. However, maintaining a policy library makes these methods inefficient spatially and temporally.

It is more efficient to use a general model for all opponent policies. Based on DQN \cite{mnih2015DQN}, DRON \cite{he2016opponent} and DPIQN \cite{hong2018DPIQN} use a secondary network which takes observations as inputs and predicts opponents’ actions. The hidden layer of this network is used by the DQN module to condition on for better policy. DRON and DPIQN are trained using the RL loss and the loss of the auxiliary task simultaneously. SOM \cite{raileanu2018modeling} uses its own policy to estimate the goals of opponents.

Representation learning is also explored for opponent modeling. These methods usually use an encoder mapping observations to the representation space. Grover \textit{et al.} [\citeyear{grover2018learning}] proposed to learn policy representations to model and distinguish agents' policies by predicting their actions and identifying them through triplet loss. In fact, the auxiliary tasks in DRON \cite{he2016opponent} and DPIQN \cite{hong2018DPIQN} can also be viewed essentially as representation learning. 

Though having achieved high performance in multi-agent tasks, many of the aforementioned methods have limitations or do not specifically consider generalization in execution. Policy reuse methods \cite{hernandez2016identifying,zheng2018deep}, SAM \cite{everett2018learning}, SOM \cite{raileanu2018modeling} and DRON \cite{he2016opponent} require opponents' observations or/and actions to do inference, which may be unrealistic in execution. DPIQN \cite{hong2018DPIQN} and Grover \textit{et al.} [\citeyear{grover2018learning}] do use local information only. However, training with action prediction as a supervision signal makes the performance rely heavily on the unseen policy to be similar enough to training opponent policies. 
Moreover, Grover \textit{et al.} [\citeyear{grover2018learning}] captures only differences between opponents' policies, but ignores similarities that we believe are important.

Our proposed MPR avoids these deficiencies by learning representations that reflect distances between opponents' policies, capturing both similarities and differences between them. Thus, MPR essentially models the policy space, so that it adapts immediately when facing unseen  opponents' policies in execution without parameter updating. Besides, the agent requires no additional information other than its own observations during execution.

\section{Preliminaries}

\subsection{Multi-Agent Environment}

We consider a setting, similar to Hong \textit{et al.} [\citeyear{hong2018DPIQN}], where there are $N+1$ \textit{independent} agents in a multi-agent environment $\mathcal{E}$: one ego agent and the other $N$ opponents. At each timestep, the ego agent selects an action $a \in \mathcal{A}$, while the other $N$ opponents' actions form a joint action $\boldsymbol{a}_o \in \boldsymbol{\mathcal{A}}_o$, where $\boldsymbol{\mathcal{A}}_o = \mathcal{A}_1\times\mathcal{A}_2\times\cdots\times\mathcal{A}_N$. The subscript $o$ denotes ``\textit{opponents}'', and $\mathcal{A}_1,\ldots,\mathcal{A}_N$ corresponds to each of the $N$ opponents' action space. The policies of the $N$ opponents form a opponent joint policy denoted by $\boldsymbol{\pi}_o(\boldsymbol{a}_o | \boldsymbol{o}_o)$, where $\boldsymbol{o}_o$ denotes the joint observation of the $N$ opponents. \textit{The policy of each opponent is sampled at the beginning of the episode and consistent within the same episode.} We define the ego agent's policy as $\pi(a|o,\boldsymbol{\pi}_o)$ to condition on $\boldsymbol{\pi}_o$. We make no assumption on agents' relations with each other: each pair of agents in $\mathcal{E}$ can be either collaborators or competitors. The reward of the ego agent at each timestep is given by a reward function $\mathcal{R}$: $r=\mathcal{R}(s,a,\boldsymbol{a}_o,s')$, and the state transition function is $\mathcal{T}(s',s,a,\boldsymbol{a}_o)=\mathrm{Pr}(s'|s,a,\boldsymbol{a}_o)$.

\subsection{Opponent Policy Space}
We notice the number of different possible policies that each opponent can take is numerous, {\em if not infinite}. For generalization to any possible policies, we consider the ``policy space'' $\Pi_i$ formed by all possible policies of opponent $i$ in $\mathcal{E}$. In each episode, opponent $i$ acts according to a policy $\pi_i$ which is sampled from a distribution $P_i$ over $\Pi_i$. Therefore, the opponent joint policy $\boldsymbol{\pi}_o$ can be viewed as sampled from the joint distribution $\boldsymbol{P}$ over the opponent joint policy space $\boldsymbol{\Pi}_o: \Pi_1\times\Pi_2\times\cdots\times\Pi_N$.

\subsection{Multi-Agent Learning Problem}

In this paper, we focus on the learning problem for the ego agent described as follows. Given environment $\mathcal{E}$ with $N+1$ agents, a training policy set $\boldsymbol{\Pi}_o^{\mathrm{train}}=\Pi_1^{\mathrm{train}}\times\Pi_2^{\mathrm{train}}\times\cdots\times\Pi_N^{\mathrm{train}}$ resembles the distribution $\boldsymbol{P}$ over $\boldsymbol{\Pi}_o$, where each $\Pi_i^{\mathrm{train}}$ resembles the distribution $P_i$ over $\Pi_i$. In each episode during training, the ego agent interacts with the other $N$ opponents with policy $\pi_1\in\Pi_1^{\mathrm{train}}, \pi_2\in\Pi_2^{\mathrm{train}}, \dots, \pi_N\in\Pi_N^{\mathrm{train}}$. These policies form an opponent joint policy $\boldsymbol{\pi}_o^{\mathrm{train}}\in\boldsymbol{\Pi}_o^{\mathrm{train}}$, on which the ego agent should condition its policy to maximize the expected return. The ego agent is tested against opponents with joint policy $\boldsymbol{\pi}_o^{\mathrm{test}}$ from the test policy set, where $\boldsymbol{\Pi}_o^{\mathrm{test}}=\Pi_1^{\mathrm{test}}\times\Pi_2^{\mathrm{test}}\times\cdots\times\Pi_N^{\mathrm{test}}$ and $\boldsymbol{\Pi}_o^{\mathrm{test}}\cap\boldsymbol{\Pi}_o^{\mathrm{train}}=\emptyset$. During test, the ego agent has to not only discriminate different opponent joint policies in $\boldsymbol{\Pi}_o$ to condition its own policy, but also calibrate the distances between policies through their representations, and thus achieve high return when facing unseen policies.

\section{Method}

\subsection{Metric Learning for Policy Representations}
Though conditioning the ego agent's policy on the opponent joint policy will be of benefit to restrain non-stationarity in our setting, it is impossible in practice because the opponent joint policy is unknown to the ego agent. Instead, we can build a representation of the opponent joint policy, on which the ego agent conditions its policy. Then, the ego agent's policy becomes $\pi(a|o,\phi(I(\boldsymbol{\pi}_o)))$, where $I(\boldsymbol{\pi}_o)$ denotes the information about the opponent joint policy that is available to the ego agent, such as the local trajectory $\tau$, and $\phi$ is a mapping from the information to the representation space.

Inspired by the theory on similarity and optimization objective setting in Ghosh \textit{et al.} [\citeyear{Ghosh2019Actionable}], MPR tries to learn policy representations that reflect the distances between policies, thus can well generalize to unseen policy with a good estimate of its relations to known policies using the policy distance measure. Such representations are generated through an encoder network $\phi$ parameterized by $\theta$, minimizing the following loss function $\mathcal{L}_{\mathrm{embed}}$:
\begin{multline}
    \mathcal{L}_{\rm embed}(\theta) = 
    \mathbb{E}_{\boldsymbol{\pi}^{i}_o, \boldsymbol{\pi}^{j}_o \sim \boldsymbol{\Pi}_o^{\mathrm{train}}}\\
    \left[\left(d_1(\phi(I(\boldsymbol{\pi}_o^i);\theta), \phi(I(\boldsymbol{\pi}_o^j);\theta))-d_2(\boldsymbol{\pi}_o^i, \boldsymbol{\pi}_o^j)\right)^2\right],
    \label{Embed-Loss}
\end{multline}
where $d_1(\cdot,\cdot)$ is a distance function between two output representations of $\phi$ network and $d_2(\cdot, \cdot)$ is a distance function between two opponent joint policies labeled by $i$ and $j$, respectively. The \textit{r.h.s} of \eqref{Embed-Loss} optimizes the encoder network $\phi$ such that the distance between the representations of each pair of $\boldsymbol{\pi}^{i}_o$ and $\boldsymbol{\pi}^{j}_o$ converges to $d_2(\boldsymbol{\pi}_o^i,\boldsymbol{\pi}_o^j)$. 

\begin{figure}[!t]
	\centering
	\includegraphics[width=0.7\columnwidth]{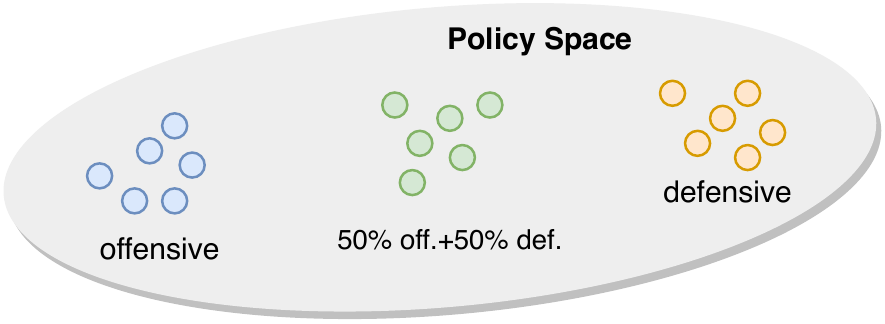}
	\caption{Intuitive illustration of policy representations reflecting their relations in policy space.}
	\label{fig:policy_space}
\end{figure}

Intuitively, for an agent in a certain environment where its policy can be offensive, defensive or halfway in the middle (50-50), the policy representations that we try to learn here, of this agent, should be able to visualize as Figure~\ref{fig:policy_space}. The policy representations should not only distinguish different policies (\textit{i.e.}, the representations of different policies gather in different clusters), but also \textit{capture their relative positions in the policy space} (\textit{e.g.}, placing the 50-50 policy in the middle of the other two), which makes our method different from common metric learning methods, for example, triplet loss used in Grover \textit{et al.} [\citeyear{grover2018learning}]. In other words, training with triplet loss maximizes distance between representations of opponents' policies with different labels, but ignores the circumstance that these opponent policies may be similar in terms of behaviour. We will show in the experiments such representations harm the performance of the ego agent.

In this paper, we use L2 distance as $d_1$ function and observation history as $I(\boldsymbol{\pi}_o)$. The encoder network is learned in a supervised way. During training, the ego agent interacts with randomly sampled opponents, and stores the whole observation history $\boldsymbol{h}=\{o_1,o_2,\dots,o_T\}$ and the opponent joint policy's label into a buffer $\mathcal{B}$. Then, pairs of observation histories are uniformly sampled from the buffer and the encoder network is optimized by
\begin{multline}
\mathcal{L}_{\rm embed}(\theta) = 
    \mathbb{E}_{<\boldsymbol{h^i},i>,<\boldsymbol{h^j},j>\sim\mathcal{B}} \\
    \left[\left(\|\phi(\boldsymbol{h^i};\theta), \phi(\boldsymbol{h^j};\theta)\|_2-d_2(\boldsymbol{\pi}_o^i, \boldsymbol{\pi}_o^j)\right)^2\right].
\label{Batch-Loss}
\end{multline}

\subsection{Policy Distance Estimation}

Now we move on to the design of $d_2$ in \eqref{Embed-Loss} and \eqref{Batch-Loss}. Intuitively, the distances between policies are naturally reflected in the differences of respective action patterns. From the ego agent's perspective, fixing its own policy and exploration, the different opponent joint policies can be captured in the different distributions of $(a, \boldsymbol{a}_o, s)$ in interactions. However, it is impractical to iterate over all possible tuples and collect enough samples to estimate this distribution, given a large state space or continuous action space. Instead, we propose to use the differences between the sampled distributions of $(a, \boldsymbol{a}_o)$ in interactions with different opponent joint policies as an approximate estimate. Given sufficient samples, the sampled frequency of $(a, \boldsymbol{a}_o)$ can be used as an approximation to the actual $\mathbb{E}_s \left[ p(a, \boldsymbol{a}_o) \right]$ which is the expected probability of $(a, \boldsymbol{a}_o)$ over all states. As we show in Appendix~\ref{sec:sampling}, this gives a good enough measurement to estimate the policy distance. Concretely, we fix the ego agent's own policy and exploration for sampling. The corresponding samples of $(a, \boldsymbol{a}_o)$ are taken by interacting with each $\boldsymbol{\pi}_o^{i}$ in training policy set $\boldsymbol{\Pi}_o^{\mathrm{train}}$ in the environment $\mathcal{E}$. 

For an environment $\mathcal{E}$ with discrete action spaces, in which $\mathcal{A}, \mathcal{A}_1, \dots, \mathcal{A}_N$ are all discrete spaces, we calculate the frequency distribution $f^{i}$ of all possible pairs of $(a, \boldsymbol{a}_o)$ using the samples taken with $\boldsymbol{\pi}_o^{i}$ as an estimate of the real probability distribution $p^{i}$. Then we use the Kullback-Leibler (KL) divergence between these distributions as a measure for the distance between the policies $\boldsymbol{\pi}_o^{i}$ and $\boldsymbol{\pi}_o^{j}$,
\begin{align}
	\label{KLdist}
	d(\boldsymbol{\pi}_o^{i},\boldsymbol{\pi}_o^{j}) & = D_{\mathrm{KL}}(p^{i}||p^{j}) + D_{\mathrm{KL}}(p^{j}||p^{i}) \nonumber \\
	& \approx D_{\mathrm{KL}}(f^{i}||f^{j}) + D_{\mathrm{KL}}(f^{j}||f^{i}).
\end{align}

\begin{figure}[!t]
	\setlength{\abovecaptionskip}{5pt}
	\centering
	\includegraphics[width=.7\columnwidth]{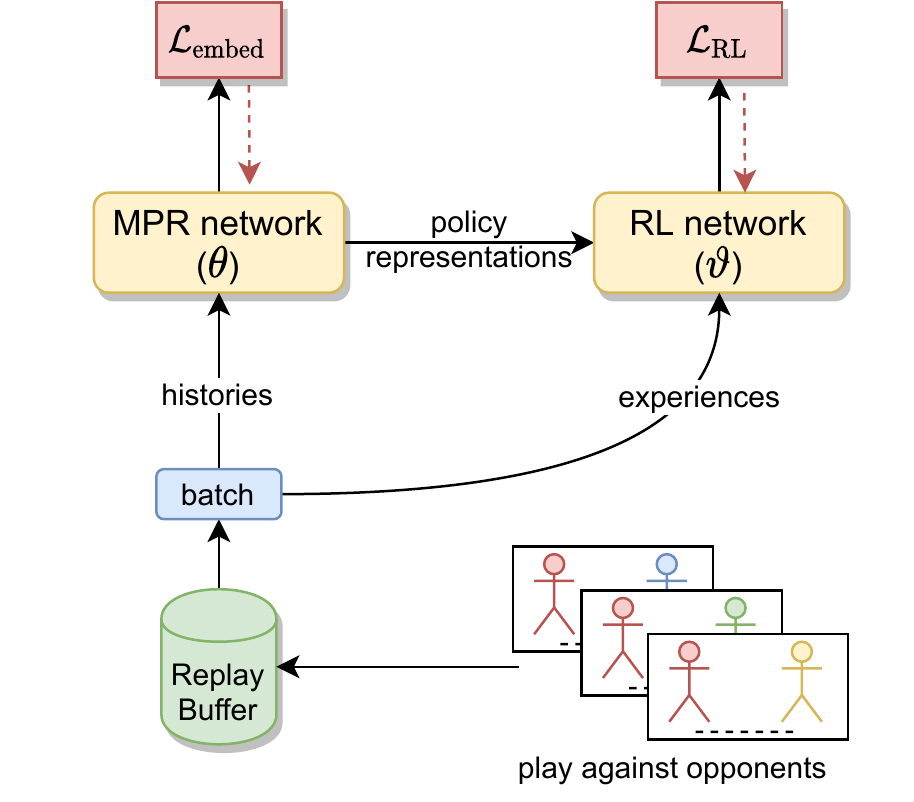}
	\caption{Illustration of MPR with RL framework.}
	\label{fig:mpr}
\end{figure}

For an environment with continuous action space, in which at least one of $\mathcal{A}, \mathcal{A}_1, \dots, \mathcal{A}_N$ is continual, we propose to use Wasserstein distance as a measurement for the distance between $\boldsymbol{\pi}_o^{i}$ and $\boldsymbol{\pi}_o^{j}$. The reason for using Wasserstein distance is that the sampled data points can be used to compute the distance directly without estimating the empirical distributions of samples, which is intractable in continuous action space,
\begin{equation}
	\label{Wdist}
	d(\boldsymbol{\pi}_o^{i},\boldsymbol{\pi}_o^{j})=W(p^{i}, p^{j}) \approx W(y^{i}, y^{j}),
\end{equation}
where $y^{i}$ is the set of $(a, \boldsymbol{a}_o)$ samples taken with $\boldsymbol{\pi}_o^{i}$. While computationally demanding in high-dimensional space, the Wasserstein distance in one-dimensional space can be computed easily. So an alternative metric, sliced Wasserstein distance \cite{Rabin2011Wasserstein}, is used to approximate the Wasserstein distance by projecting the raw $m$-dimensional data points into one-dimensional space and computing one-dimensional Wasserstein distance:
\begin{equation}
	SW(X,Y) = \int_{\sigma\in \mathbb{S}^{m-1}} W(\sigma^TX, \sigma^TY)d\sigma
\end{equation}
where $\mathbb{S}^{m-1}$ is the unit sphere in $m$-dimensional space. In practice, the sliced Wasserstein distance is usually approximated by summation over randomly projections \cite{DBLP:conf/cvpr/DeshpandeZS18}:
\begin{equation}
	\tilde{SW}(X,Y) = \frac{1}{|\Omega|} \sum_{\sigma\in\Omega} W(\sigma^TX, \sigma^TY)
\end{equation}
where $\Omega$ is the set of randomly generated projections. We use the approximated sliced Wasserstein distance in practice to reduce the computational overhead.

\begin{algorithm}[t]
	\caption{Joint Training of MPR and RL} 
	\begin{small}
		\label{alg:joint-training}
		\begin{algorithmic}[1]
			\REQUIRE Training policy set $\boldsymbol{\Pi}^{\mathrm{train}}_o = \left\{\boldsymbol{\pi}_o^1, \boldsymbol{\pi}^2_o, \dots, \boldsymbol{\pi}^{\mathcal{N}}_o\right\}$
			\STATE Initialize $\mathcal{E}$, replay buffer $\mathcal{B}$, MPR network parameters $\theta$, RL network parameters $\vartheta$
			\STATE Initialize joint-action distributions $\mathcal{D}_1,\mathcal{D}_2,\dots,\mathcal{D}_\mathcal{N}$ 
			\FOR{$\boldsymbol{\pi}^k_o \in \boldsymbol{\Pi}^{\mathrm{train}}_o$}
			\STATE Set opponent joint policy to $\boldsymbol{\pi}_o^k$
			\STATE Roll out $num\_sample$ episodes $\boldsymbol{e}$
			\STATE Update $\mathcal{D}_k$ with all $(a, \boldsymbol{a}_o)$ pairs from $\boldsymbol{e}$
			\ENDFOR
			\STATE Calculate $d\left(\boldsymbol{\pi}^i_o, \boldsymbol{\pi}^j_o\right)$ by \eqref{KLdist} or \eqref{Wdist} for all $(i, j)$ using $\mathcal{D}_i, \mathcal{D}_j$
			\FOR{iteration $t=1$ to $T$}
			\STATE Randomly select a $\boldsymbol{\pi}_o$ from $\boldsymbol{\Pi}_o^{\rm train}$
			\STATE Set opponent joint policy to $\boldsymbol{\pi}_o^k$
			\STATE Rollout $num\_collect$ episodes $\boldsymbol{e}$ and store $\boldsymbol{e}$ into $\mathcal{B}$
			\STATE Sample a batch from $\mathcal{B}$ and update $\vartheta$ by $\mathcal{L}_{\rm RL}$
			\STATE Sample a batch from $\mathcal{B}$ and update $\theta$ by $\mathcal{L}_{\rm embed}$
			\IF{MPR-RS \AND $t\bmod resample\_period = 0$}
			\STATE Repeat step 3 - 8
			\ENDIF
			\ENDFOR
		\end{algorithmic}
	\end{small}
\end{algorithm}

\subsection{Metric Policy Representations with RL}

\begin{figure*}[t]
	\centering
	\includegraphics[width=1.6\columnwidth]{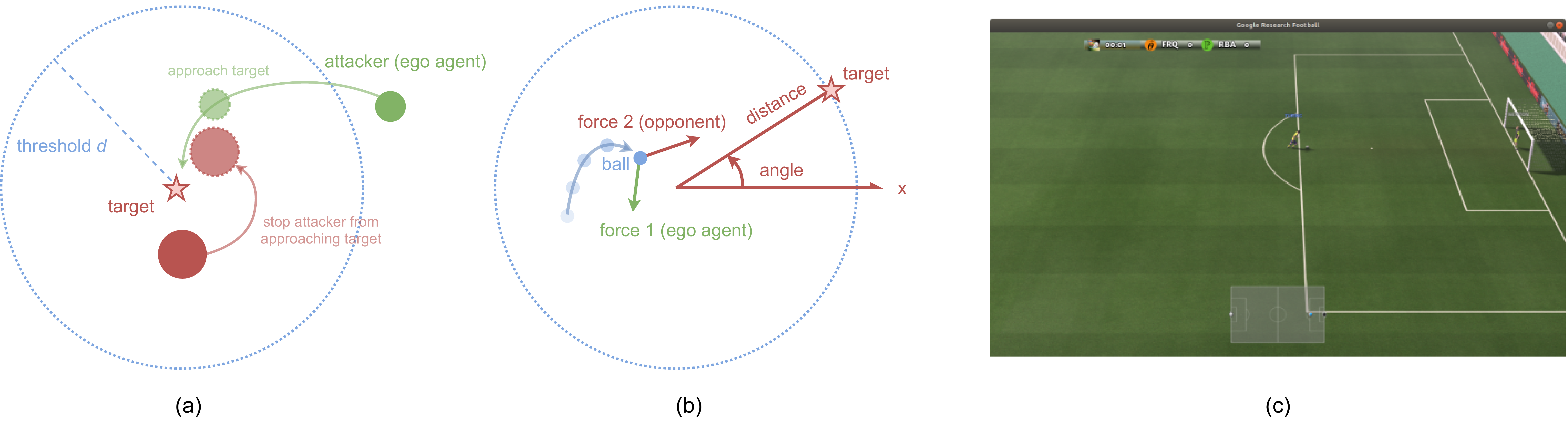}
	\caption{Illustration of (a) \texttt{Push}, (b) \texttt{Keep}, (c) \texttt{1v1 Football}}
	\label{fig:envs}
\end{figure*}

Figure~\ref{fig:mpr} illustrates a general framework that combines MPR with RL algorithms. The histories and experiences are stored together in the replay buffer. The MPR network $\phi(\boldsymbol{h};\theta)$, \textit{i.e.}, the encoder network described above, takes histories from replay buffer as input to generate representations of opponent joint policies in the corresponding episodes, and optimizes its parameters $\theta$ to minimize $\mathcal{L}_{\mathrm{embed}}$. The representation of current opponent joint policy, generated by taking as input the history so far in current episode, is fed into the RL network as an additional input, making the ego agent's policy condition on opponents' policies $\boldsymbol{\pi}_o$ approximately. The RL module optimizes its parameters $\vartheta$ to minimize its own loss $\mathcal{L}_{\mathrm{RL}}$ conditioned on the representations. We optimize $\theta$ and $\vartheta$ using different optimizers and do not back-propagate the RL loss to MPR network.

The training process is described in Algorithm~\ref{alg:joint-training}. The empirical joint action distribution can be sampled against opponents at the beginning of the training, or by periodically sampling along with training. Two options are denoted as MPR-NoRS and MPR-RS (where RS stands for \textit{Re-Sample}), respectively. MPR-RS continues to obtain current joint action distributions under the ego agent's updated policy, but also produces moving targets for $\mathcal{L}_{\mathrm{embed}}$. MPR-NoRS has a fixed optimization target for better convergence, but the calculated distances might not give an useful estimation in later period of training. In practice, as we will show in experiments, the choice of the two is made based on the environment settings. 

Note that only opponents' action $\boldsymbol{a}_o$ and their joint policy label $i$ are required for obtaining the joint-action distributions and estimating policy distances during training. In execution, only local observed history $\boldsymbol{h}$ are needed to generate a good policy against diverse opponents, which is practical in most scenarios.

\section{Experiments}

We evaluate MPR in three multi-agent environments: \texttt{Push}, \texttt{Keep} and \texttt{1v1 Football}.

\subsection{Push}
\label{sec:push-exp}

\subsubsection*{Task and Setting}

Our \texttt{Push} environment is modified based on the original \texttt{simple-push} scenario of Multi-Agent Particle Environment \cite{mordatch2017emergence,lowe2017multi}. As shown in Figure~\ref{fig:envs}(a), one landmark (``target") fixed at origin $(0,0)$, and two competitive agents---an ego agent (``attacker") which tries to approach and touch the target, and a defensive opponent (``defender") which tries to stop the attacker from approaching the target and push it away. Each agent has a \textit{discrete action space} consists of 5 actions and corresponds to applying a zero force, or a unit force on four directions.

The defender's policy is rule-based. Each timestep, the defender calculates the attacker's distance to the target using its observation. If the distance is larger than a threshold parameter $d$, the defender moves towards the target, otherwise it moves towards the attacker. 



In this experiment, the training policy set consists of 4 policies with different threshold parameter $d$ for the defender: $d=0.1, 0.3, 0.75, 1.0$. The testing policy is set with $d=0.5$. We combine our MPR method with DQN \cite{mnih2015DQN}, and compare the performance when interacting with the test opponent with DPIQN \cite{hong2018DPIQN} which uses the same architecture as DQN+MPR but trains the MPR network with cross-entropy loss to predict the opponents' actions and with RL loss gradients back-propagated to the encoder, DQN+Triplet which uses the same architecture as DQN+MPR but trains the MPR network with triplet loss
and vanilla DQN without the encoder network of MPR. More details about the environment, networks and hyperparameters are available in Appendix~\ref{sec:push_detail}.

\begin{figure}[t]
	\setlength{\abovecaptionskip}{3pt}
	\centering
	\includegraphics[width=1.0\columnwidth]{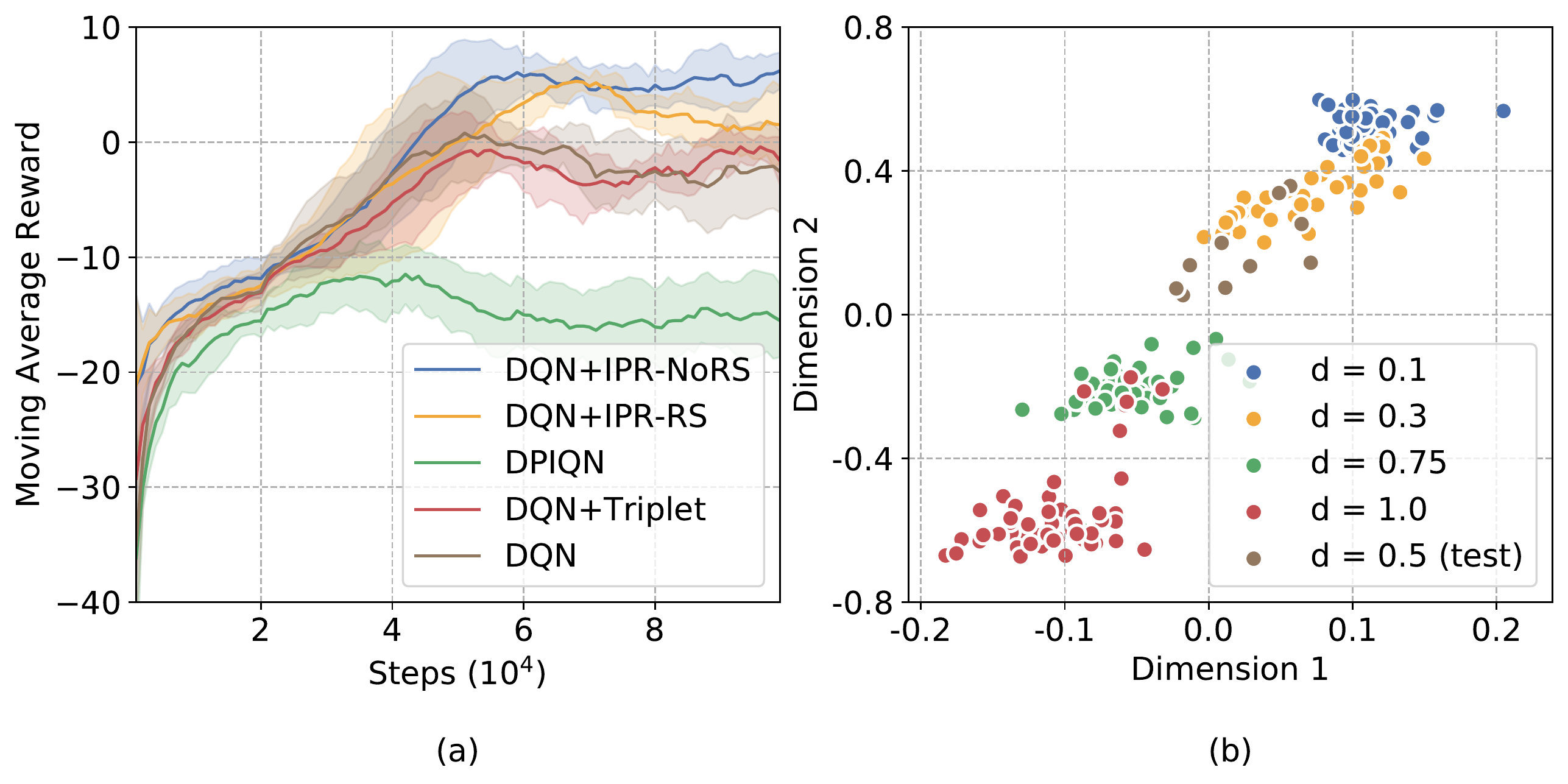}
	\caption{(a) Moving average of test rewards in \texttt{Push}. Each curve corresponds to the mean value of 5 trials with different random seeds, and shaded regions indicate standard deviation. (b) Opponent policy representations generated by MPR network in last 10 timesteps in an episode on \texttt{Push}. Dimension reduced by multidimensional scaling (MDS) to 2.}
	\label{fig:push_result}
\end{figure}

\subsubsection*{Quantitive Results} Figure~\ref{fig:push_result}(a) shows the moving average reward when testing against the test policy ($d=0.5$) of our method and the baselines. The tests are run every 1000 steps during training, and the moving average reward is taken by the mean of recent 20 tests. As the result shows, DQN+MPR-NoRS and DQN+MPR-RS outperform the compared methods when facing the unseen policy, thus proving the effectiveness of our proposed method. MPR-NoRS performs sightly better than MPR-RS in this environment, which might be the result of the moving target problem and also of the learned policy limits the re-sampling from capturing as much information of different policies as initial random sampling.

Regarding the other compared methods, we infer that because of the threshold settings, DPIQN can hardly learn useful predictions on opponent's actions through observed history, from which can hardly deduce which of the two acting patterns the opponent will act on. DQN essentially views the different policies as one, and triplet loss maximizes the distance between policy representations regardless of their relations. Thus all compared methods perform worse than MPR.

\subsubsection*{Visualization of Learned Representations}

In Figure~\ref{fig:push_result}(b), we visualize the learned policy representations (dimension reduced by MDS from 32 to 2) output by the encoder of one learned DQN+MPR-NoRS model on \texttt{Push}. The training policy representations ($d=0.1, 0.3, 0.75, 1.0$) are generated by the MPR encoder network from histories that are randomly taken from the final replay buffer. The test policy representations ($d=0.5$) are generated in actual test. Each shown representation corresponds to one of the last 10 timesteps in an episode. The result shows that the learned representation, including training and testing policies, can reflect to their relations (essentially the relations between different $d$), thus validate our method's hypothesis of policy representations reflecting the policy space, resembling the expected result in Figure~\ref{fig:policy_space}.

\subsection{Keep}
 
\subsubsection*{Task and Setting}
In order to verify the effectiveness of our method in \textit{continuous action space}, we implement a simple two-agent environment named \texttt{Keep}. At the beginning of each episode, a ball is initialized around the origin $(0,0)$. As illustrated in Figure~\ref{fig:envs}(b), the goal of the ego agent is to keep the ball close to the origin, while the opponent tries to pull the ball away from the origin. The opponent’s policy is rule-based and can be described as a triple (\textit{angle}, \textit{distance}, \textit{force}). For example, $(45.0, 1.0, 0.5)$ means that the target is located at a distance of 1.0 from the origin and its angle from the x-axis is $45$ degrees, same as polar coordinate, and the opponent will pull the ball towards the target with a force of $0.5$.  The agent’s observation includes the coordinate and velocity of the ball, and the action is the magnitude and direction of the force exerted on the ball. Each timestep, the ego agent receives a negative reward of the distance between the ball and the origin.

In this experiment, the training policy set contains 4 different opponent policies: $(45.0, 1.0, 0.5)$, $(170.0, 2.0, 1.0)$, $(-90.0, 1.5, 0.7)$, and $(0.0, 1.0, 0.3)$. In test phase, the opponent policy is generated randomly at the beginning of each episode, where the ranges of angle, distance and force are $[-180, 180)$, $[0, 1)$ and $[0.2, 1.7)$, respectively. We combine our method with PPO \cite{DBLP:journals/corr/SchulmanWDRK17} and compare our method (PPO+MPR-NoRS and PPO+MPR-RS) with PPO+ActPred (training the MPR network by predicting opponents' actions, like DPIQN), PPO+Triplet, and vanilla PPO. More details about the networks and hyperparameters are available in Appendix~\ref{sec:keep_detail}.

\subsubsection*{Quantitive Results}

\begin{figure}[t]
	\setlength{\abovecaptionskip}{3pt}
	\centering
	\includegraphics[width=1.0\columnwidth]{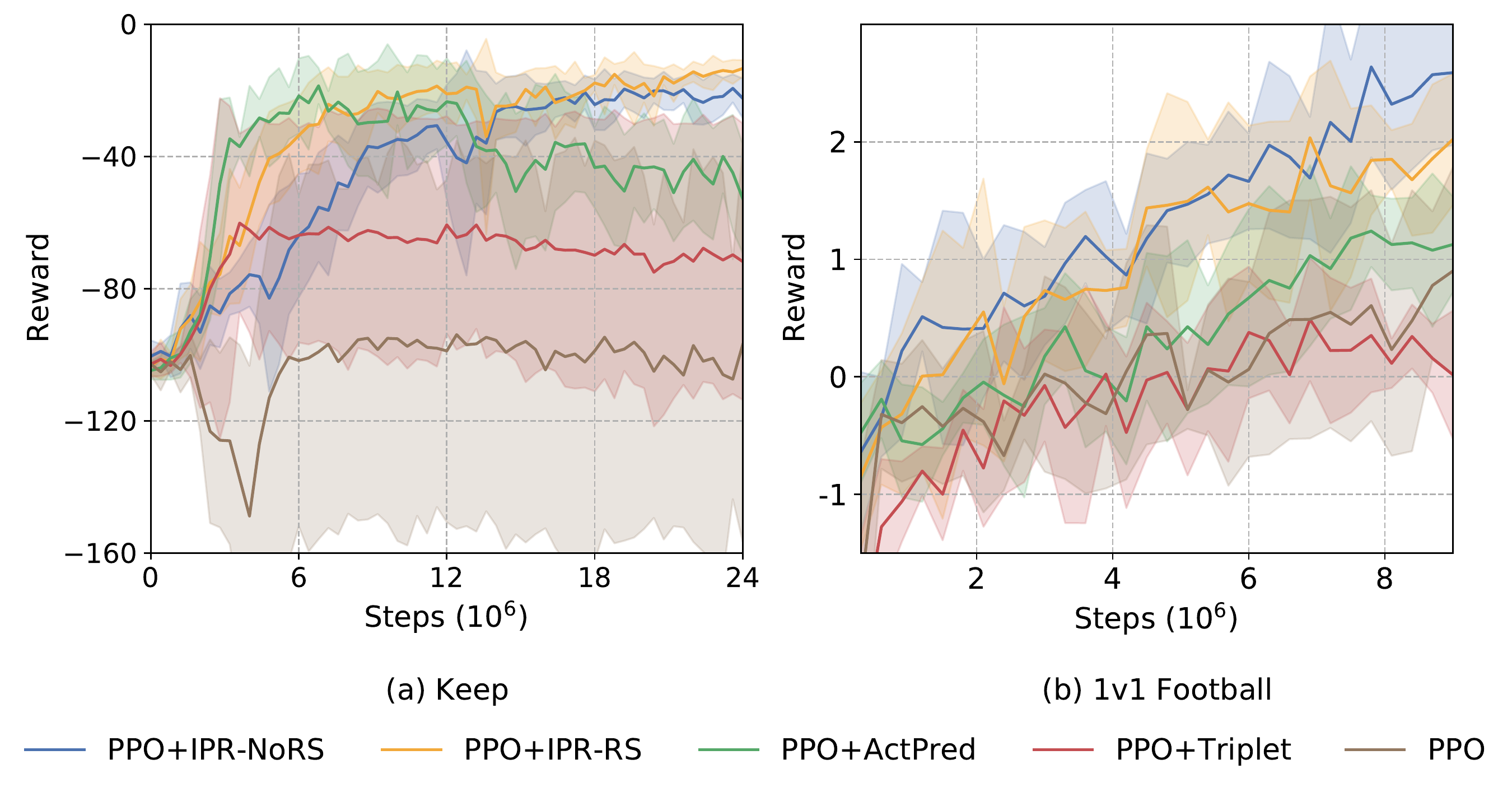}
	\caption{(a) Average rewards against testing opponents on \texttt{Keep}. Each curve shows the mean reward of 5 trials with different random seeds, and shaded regions indicate standard deviation. (b) Average rewards against testing opponents on \texttt{1v1 Football}. Each curve shows the mean reward of 5 trials with different random seeds, and shaded regions indicate standard deviation.}
	\label{fig:ppo_result}
\end{figure}

Figure~\ref{fig:ppo_result}(a) shows the average reward against the random testing opponent throughout training phase. The result suggests that PPO+MPR-NoRS and PPO+MPR-RS outperform other methods. Although PPO+ActPred achieves almost the same reward as the former two early in the training, it shows high variance and its reward declines with training. PPO+MPR-NoRS and PPO+MPR-RS are more stable and continuously improve in the later stage of training, which implies our method's effectiveness and better generalization to unseen opponents also in continuous action space. PPO and PPO+Triplet fail to learn an efficient policy against unseen opponents and show high variance. In addition, MPR-RS performs sightly better than MPR-NoRS in this environment, possibly because the initial sampling has interference from action pairs that sampled when the ball is far from the origin, while the later learned policy avoids such interference so different opponents' policies show more individual characteristics. Analysis of learned representations is available in Appendix~\ref{sec:keep_emb}.

\subsection{1v1 Football}

\subsubsection*{Task and Setting}

\texttt{1v1 Football} is a two-agent environment implemented on Google Research Football \cite{kurach2020google} consists of a shooter and a goalkeeper, as shown in Figure~\ref{fig:envs}(c). We build an environment wrapper, where each agent can observe positions and directions of itself, the opponent and the ball and has a discrete action space. The ego agent controls the goalkeeper and the opponent controls the shooter. We pre-trained 10 different LSTM policies for the opponent, labeled by $\pi_o^0,\dots,\pi_o^9$. Compared with the former two experiments, the environment and the opponents become much more complicated.

In this experiment, the training opponent policy set $\boldsymbol{\Pi}_o^{\rm train} = \{\pi_o^0, \pi_o^0, \pi_o^1, \pi_o^2\}$. Since we cannot control the similarity between network-based polices the way as easy as between rule-based policies, we put two same policies ($\pi_o^0$) in our training set to introduce similarity. The remaining policies ranging from $\pi_o^3$ to $\pi_o^9$ compose the test set. We combine our method with PPO and compare the performance with PPO+ActPred , PPO+Triplet, and vanilla PPO. More details about the environment, networks and hyperparameters are available in Appendix~\ref{sec:football_detail}.

\subsubsection*{Quantitive Results}

Figure~\ref{fig:ppo_result}(b) shows the average reward against the testing opponents throughout training phase. The result suggests that PPO+MPR-NoRS and PPO+MPR-RS greatly outperform other methods, proving the effectiveness of our proposed method in complicated environments. We found that the pre-trained LSTM policies are highly stochastic, making the goalkeeper difficult to predict the shooter's action. Thus, PPO+ActPred performs worse than PPO+MPR-NoRS and PPO+MPR-RS. Metric learning can solve this problem by enforcing encoder focus on temporally global information over the whole episode to generate more stable representations, keeping our MPR effective. Though PPO+Triplet are also based on metric learning, it fails by maximizing the distance between representations of similar shooter policies, \textit{i.e.}, two $\pi_o^0$ in the training set.
\section{Conclusion}


In this paper, we propose a general framework, MPR, that learns policy representations that reflect distance between opponent joint policies via joint-action distributions in multi-agent scenarios. Combining with existing RL algorithms, the policy takes actions conditioned on the learned policy representations of opponents. Through experiments, we demonstrate that the proposed framework can generalize better to unseen policies than existing methods, and the visualizations of the learned policy representations verify they can reflect the distance between policies in the policy space.

\begin{footnotesize}
\bibliographystyle{named}
\bibliography{ref}
\end{footnotesize}

\clearpage

\appendix

\section{Empirical Joint Action Distribution Sampling Result}
\label{sec:sampling}

\begin{figure}[t]
	\centering
	\includegraphics[width=0.9\columnwidth]{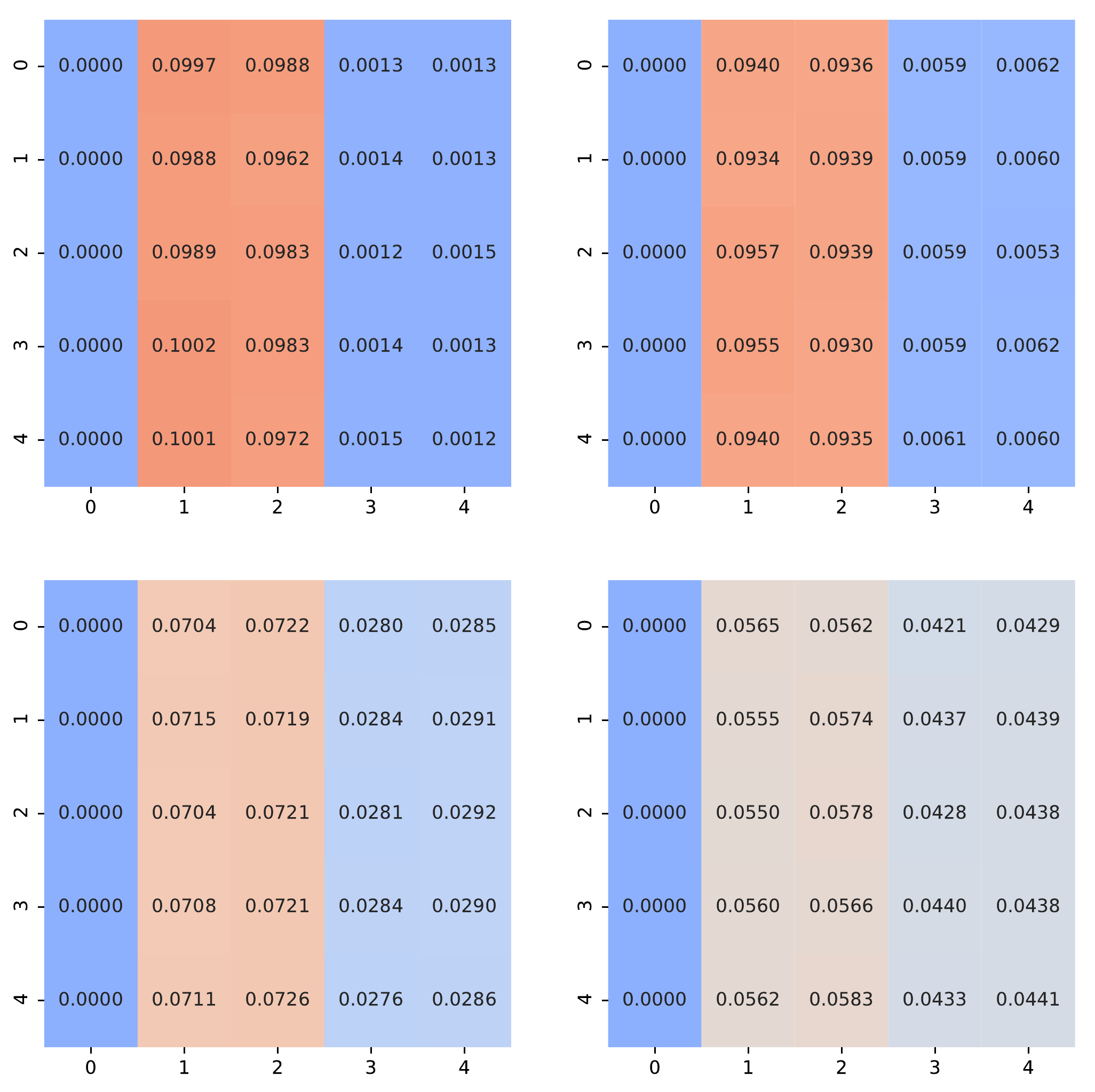}
	\caption{Heatmaps of the initial sampling results against the four training opponent policies under one random seed in \texttt{Push}, where x-axis indicates the opponent's action and y-axis the (random) agent's. Colors and values indicate the frequency of the corresponding joint-action pairs. Action 0 exerts zero force, and actions 1-4 each exerts unit force towards one of the four directions.}
	\label{fig:act_arr}
\end{figure}

As described in the main text, we empirically use the sampled frequency of $(a, \boldsymbol{a}_o)$ as an approximation to the actual $\mathbb{E}_s \left[ p(a, \boldsymbol{a}_o) \right]$. Figure~\ref{fig:act_arr} illustrates this estimation. The four heatmaps show the initial sampling results using random policy against the four training opponent policies $d = 0.1, 0.3, 0.75, 1.0$ in order, and the colors and values on each square of the heatmaps indicate the frequency of the corresponding joint-action pair composed by the agent's action (y-axis) and the opponent's (x-axis). It is easy to see that the frequency of every joint-action pair (except for the opponent taking 0) increases or decreases monotonically across the four heatmaps, which corresponds to the orderly change of the parameter $d$. This monotonic change is essentially caused by the different distributions $p(a, \boldsymbol{a}_o)$ under different opponent policies. When $d=1.0$, the opponent is more aggressive and chases after the ego agent everywhere, while when $d=0.1$, the opponent stays around the origin for the most time. The different acting patterns cause the difference in $p(a, \boldsymbol{a}_o)$ that is reflected in $(a, \boldsymbol{a}_o)$ frequencies, thus allow our sampling to have a good estimation.

\section{Additional Details on \texttt{Push} Experiments}
\label{sec:push_detail}

The defender has a larger size and mass but a smaller acceleration and maximum velocity than the attacker. At each timestep, the attacker receives a negative reward of its distance to the landmark, $+2$ reward if it touches the landmark, and $-2$ reward if it is collided with the defender. The ego agent's observation at each timestep consists of the its relative positions with target and defender, and its own velocity. The opponent's velocity is unknown to the ego agent, thus making \texttt{Push} partially observable.  

For all the methods, the Q-network consists of 4 fully connected layers, with 128 hidden units and ReLU activation function in each layer. For all the methods except DQN, the encoder network consists of a 2-layer LSTM and an embedding layer that outputs the 32-dimensional representation. The number of hidden units in each layer is 128. The output 32-dimensional representation is then feed to the Q-network where it is concatenated with the output of the first layer of the Q-network. The concatenation is then forwarded through the rest fully connected layers. 

For training, the learning rates of Q-network and encoder are both set to 1e-3. Q-network and encoder train simultaneously with shared batch with size 64. The discount factor is 0.99. In the MPR experiments, 200 episodes are collected for the initial random sampling of joint action distributions.

\section{Additional Details on \texttt{Keep} Experiments}
\label{sec:keep_detail}

For all the methods, the policy network consists of 3 fully connected layers, with 32, 64 and 32 hidden units, respectively, and Tanh activation function in each layer. Value network shares the same architecture except the last output layer. For all the methods except vanilla PPO, the encoder network consists of a 1-layer GRU with hidden size 32 and a fully connected layers with 32 hidden units and Tanh activation function. The output 32-dimensional representation is concatenated with the output of the first layer in policy/value network. The concatenation is then forwarded through the rest fully connected layers. 

For training, the learning rates of all network are set to 3e-4. Unlike DQN, PPO is an on-policy algorithm, so we use two buffers separately. After each collection of 64 episodes, policy network and value network update using PPO with batch size 1024 and repeat 80 times, and encoder update 15 times with batch size 20. The discount factor is 0.99 and the GAE lambda is 0.95.

\section{Additional Details on \texttt{1v1 Football} Experiments}
\label{sec:football_detail}

For shooter, there are 11 alternative actions including idle, move to one of 8 directions, shoot and release direction. For goalkeeper, there are 12 alternative actions including idle, move to one of 8 directions, long pass, release direction and sliding. The length of each episode is set to 150. In one episode, there are multiple rounds. At the beginning of a round, the shooter is reset to the midpoint of the edge of penalty area and the goalkeeper is reset to the middle of the goal, as shown in Figure~\ref{fig:envs}(c). A round ends when the shooter scores, the goalkeeper catches the ball or timestep reaches 30. The goalkeeper receives $-1$ reward if the shooter scores and $+1$ otherwise at the end of a round. Because of multiple rounds, the ego agent, \textit{i.e} the goalkeeper, has the opportunity to observe the shooter and get its policy representation, and thus adopt the corresponding goalkeeping policy in the following rounds.

Shooter policy consists of a RNN layer with hidden size 64 and a fully connected layer with hidden size 32 and Tanh activation function. We use PPO to pre-train 10 shooter policies with the other 10 goalkeeper policies simultaneously. The pre-trained goalkeeper policies have nothing to do with the ego agent in the experiment.

All networks in this experiment have the same architecture as that in \texttt{Keep} experiment. For training, the learning rates of all network are set to 3e-4. After each collection of 16 episodes, policy network and value network update using PPO with batch size 64 and repeat 10 times, and encoder update 6 times with batch size 10. The discount factor is 0.99 and the GAE lambda is 0.95.

\section{Visualization of Learned Representations in \texttt{Keep}}
\label{sec:keep_emb}

\begin{figure}[t]
	\centering
	\includegraphics[width=0.6\columnwidth]{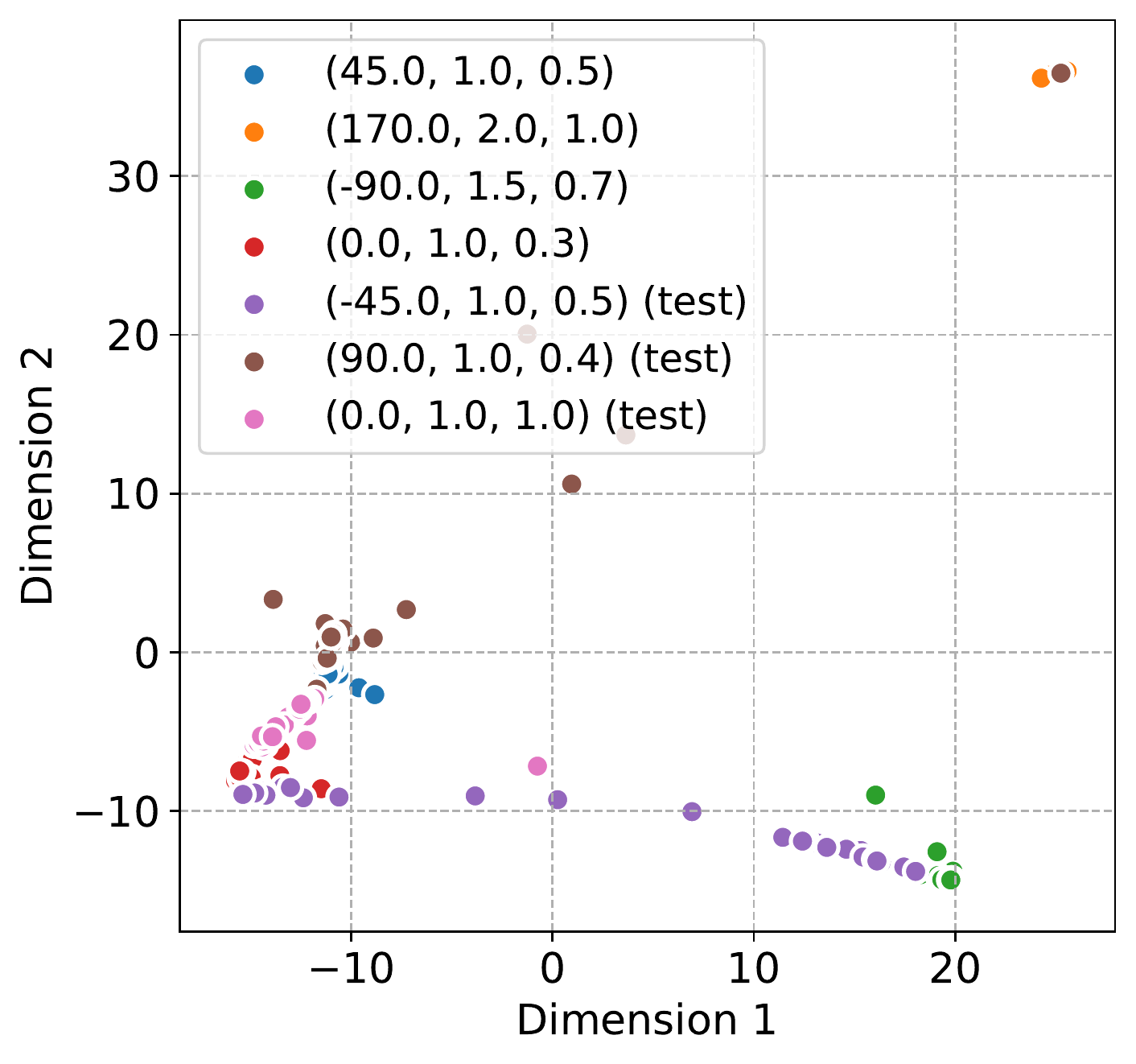}
	\caption{Opponent policy representations generated by MPR network on \texttt{Keep}, including four training opponents and three unseen test opponents. Dimension reduced by MDS to 2.}
	\label{fig:keep_emb}
\end{figure}

To demonstrate the interpretability of the representations learned by PPO+MPR-RS in \texttt{Keep}, we select three unseen opponents for test: $(-45.0, 1.0, 0.5)$, $(90.0, 1.0, 0.4)$ and $(0.0, 1.0, 1.0)$. MDS on the representations is shown as Figure~\ref{fig:keep_emb}. Each point represents the average representation over one episode. Among training opponent policies, the distance between the representations of $(45.0, 1.0, 0.5)$ and $(0.0, 1.0, 0.3)$ is the smallest, because their \textit{angle} parameters are the closest. As for unseen opponents, we can see the representations of $(-45.0, 1.0, 0.5)$ are located near $(-90.0, 1.5, 0.7)$ and $(0.0, 1.0, 0.3)$, which is consistent with the physical meaning that $-45.0$ is greater than $-90.0$ but less than $0.0$ and $0.5$ is less than $0.7$ but greater than $0.3$, even though the ego agent has not met $(-45.0, 1.0, 0.5)$ during training. Other meaningful points include the representations of $(0.0, 1.0, 1.0)$ are quite close to $(0.0, 1.0, 0.3)$, and $(90.0, 1.0, 0.4)$ are near $(45.0, 1.0, 0.5)$ and $(170, 2.0, 1.0)$.

\section{Future Work}

Though we show performance improvement and well-learned representation in multi-agent environments, estimating policy distance using joint action distributions is not a perfect method. For example, temporal information is lost in the joint action distribution, which may be a critical feature of the policy in some scenarios. We leave finding more accurate metric as future work.

\end{document}